\documentclass[runningheads]{llncs}

\usepackage{eccv}

\usepackage{eccvabbrv}

\usepackage{graphicx}
\usepackage{booktabs}
\usepackage{wrapfig}

\usepackage{multirow}
\usepackage{colortbl}
\usepackage{bm}
\usepackage{multirow}
\usepackage{xcolor}
\usepackage{colortbl}
\usepackage{amsmath,bm}
\usepackage[percent]{overpic}

\definecolor{firstcolor}{HTML}{BDE6CD}
\definecolor{secondcolor}{HTML}{E2EEBC}
\definecolor{thirdcolor}{HTML}{FFF8C5}

\newcommand{\fst}[1]{\cellcolor{firstcolor}\bfseries #1}
\newcommand{\snd}[1]{\cellcolor{secondcolor}#1}
\newcommand{\trd}[1]{\cellcolor{thirdcolor}#1}

\newcommand{\method}{MAGiSt3R}
\newcommand{\ltog}{MAGMA}

\usepackage[accsupp]{axessibility}  

\usepackage[pagebackref,breaklinks,colorlinks,citecolor=eccvblue]{hyperref}

\usepackage{orcidlink}

\begin{document}

\title{\method{}: Multi-Agent Feed-forward 3D Reconstruction from Monocular RGB Videos}

\titlerunning{\method{}}

\author{
	Ziren Gong$^{1}$\orcidlink{0000-0003-0093-835X} \quad\quad\quad Xiaohan Li$^{2}$\orcidlink{0000-0003-3109-9413} \quad\quad\quad Fabio Tosi$^{1}$\orcidlink{0000-0002-6276-5282} \quad\quad\quad Ninghui Xu$^{3}$\orcidlink{0000-0002-0652-2788} \\ Stefano Mattoccia$^{1}$\orcidlink{0000-0002-3681-7704} \quad\quad\quad Jianfei Cai$^{4}$\orcidlink{0000-0002-9444-3763} \quad\quad\quad Matteo Poggi$^{1}$\orcidlink{0000-0002-3337-2236} 
}

\authorrunning{Gong et al.}

\institute{University of Bologna, Italy \and
Faculty of Dentistry, The University of Hong Kong, China \and
School of Instrument Science and Engineering, Southeast University, China \and Monash University, Australia \\
\texttt{Project page:} \url{https://zorangong.github.io/magist3r_page/}
}

{
\maketitle
\begin{center} 
    \centering
    \begin{overpic}[width=\linewidth,scale=1.00, clip,trim=0cm 0cm 0cm 0cm,]{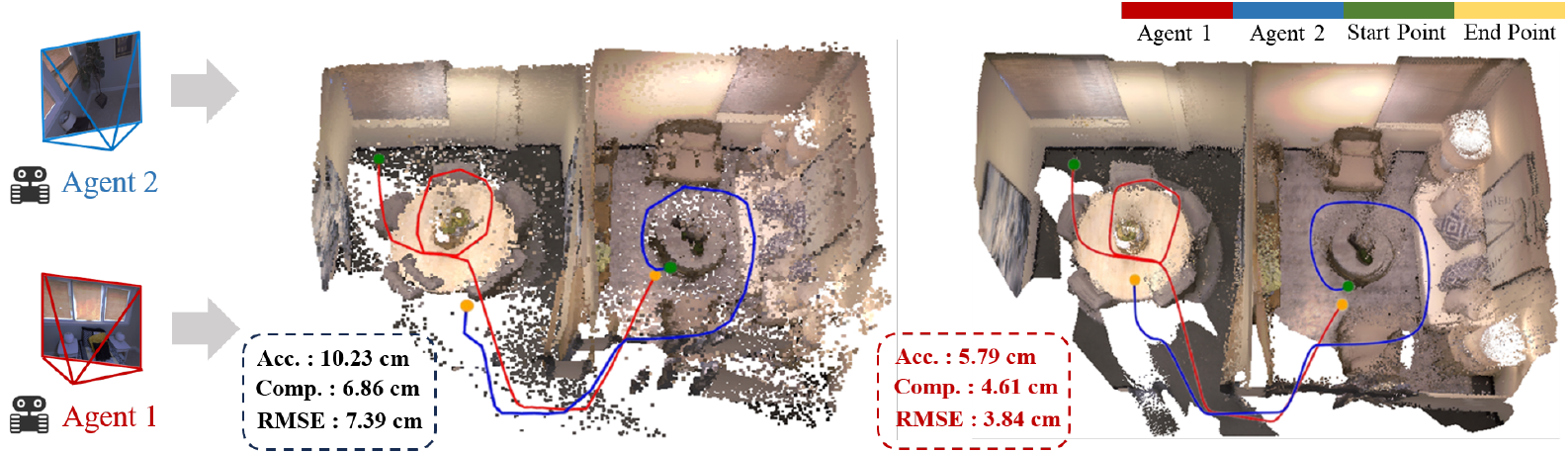}
    \put (23,-2.7) {\small VGGT-SLAM \cite{maggio2025vggt} + RICP}
    \put (65,-2.7) {\small \textcolor{purple}{\textbf{\method{} (Ours) }}}
    \end{overpic}\vspace{0.2cm}
    \captionof{figure}{\textbf{\method{} -- multi-agent 3D reconstruction in action.} 
    We showcase the mapping and tracking results of our method against VGGT-SLAM~\cite{maggio2025vggt}+RICP (RANSAC+ICP) on ReplicaMultiagent \emph{Apart 2}. Our framework produces more accurate and consistent reconstruction, along with precise camera trajectories.
    }
    \label{fig:teaser}
\end{center}
}

\begin{abstract}
This paper presents \method, a multi-agent 3D reconstruction framework performing reconstruction and camera tracking for monocular RGB videos at almost 10 FPS.
\method{} relies on a feed-forward model from the 3R family to process RGB videos and regress local point maps, and on a merging model, \ltog{}, that combines local maps at both intra-agent and inter-agent levels to obtain the final, global point map.
Furthermore, \method{} performs pose graph optimization to mitigate cumulative camera drift occurring along the feed-forward pipeline. We evaluate \method{} on both synthetic and real-world datasets, demonstrating its superior reconstruction and camera tracking accuracy compared to state-of-the-art feed-forward approaches.
\end{abstract}    
\section{Introduction}
\label{sec:intro}

Multi-agent systems have emerged as a powerful paradigm in the broad field of artificial intelligence, enabling multiple autonomous entities to collaborate to achieve complex goals. Recent advances in agentic AI have further emphasized the importance of systems that, in addition to perceiving and reasoning about the observed environment independently, can also proactively coordinate, share information, and act toward common objectives. This paradigm has also reached the intersection of robotics and computer vision, with the 3D reconstruction task, at the core of autonomous navigation and other higher-level applications, being lifted into a collaborative process, with multiple agents navigating independently in the environment and contributing to the mapping task \cite{hu2023cp,deng2025mne,yugay2025magic}. 

Through the decades, 3D reconstruction has been approached in different flavors, mainly categorized into two families: offline approaches, such as Structure-from-Motion (SfM) \cite{lindenberger2021pixel, liu20233d, schonberger2016structure, snavely2006photo} and Multi-View Stereo (MVS), assuming the full availability of the images to process in advance; and online methods, such as Simultaneous Localization and Mapping (SLAM) systems \cite{campos2021orb, engel2014lsd, teed2021droid}, assuming image streams as the input.
The latter is the one most suitable for real systems where, as soon as new images are collected, a prompt reaction is needed. 

SLAM paradigm has faced a revolution \cite{tosi2024nerfs} recently. Indeed, the adoption of Neural Radiance Fields (NeRF \cite{mildenhall2020nerf}) at first and 3D Gaussian Splatting (3DGS \cite{kerbl20233d}) later as main engines for the mapping process has empowered SLAM systems to yield dense 3D reconstructions and possibly render the scene from novel, arbitrary viewpoints. This, however, came with significant memory requirements and runtime constraints, due to the need for performing per-scene optimization at deployment. 
On top of these advances, multi-agent SLAM systems \cite{deng2025mne,yugay2025magic} have opened up new possibilities in spatial AI applications, possibly accelerating the reconstruction process while enforcing geometric consistency on the collaboratively reconstructed global map. 
Specifically, multi-agent systems built on top of NeRF \cite{deng2025mne} and 3DGS \cite{yugay2025magic} engines have been proposed, respectively, yet inheriting the main limitations of NeRF and 3DGS, thus being i) scene-specific, since requiring the optimization process to take place directly during navigation, and ii) often unable to process under real-time constraints due to their iterative parameter updates.

In contrast, the advent of 3D vision foundation models, often referred to "3R" models \cite{wang2024dust3r}, has ignited a new interest in developing feed-forward 3D reconstruction systems \cite{liu2025slam3r,maggio2025vggt,gong2026ov3r}, complementing the latest NeRF/3DGS-based SLAM pipelines. On the one hand, these models can run on any unconstrained RGB video, not requiring any additional information such as depth -- even with unknown camera parameters -- and are no longer optimized over single sequences, thus allowing for much faster inference. On the other hand, 3R models processing video streams often lack global optimization, thus being prone to trajectory drift. 
In light of this, deploying 3R models in a multi-agent framework has the potential to exploit their strengths at best, while overcoming their main weaknesses related to camera drifting over long video sequences. However, no multi-agent feed-forward 3D reconstruction system has been developed to date.

Driven by these arguments, we introduce \method{}, the first multi-agent feed-forward 3D reconstruction framework for monocular RGB videos.
Built on the latest advances in 3R models, \method{} globally and incrementally reconstructs scenes where multiple agents navigate concurrently. This is achieved according to the following three main steps: i) \textbf{Intra-agent reconstruction}: each agent processes the input RGB sequence in sub-maps through a 3R backbone, then merges the sub-maps incrementally into a geometrically consistent one; ii) \textbf{Inter-agent alignment}: \method{} leverages a loop closure mechanism to detect overlapping regions between multiple agents. Once a loop is detected, the point maps from multiple agents are combined into the final global map;
iii) \textbf{Pose graph optimization}: tracked poses are optimized after merging sub-maps at the intra-agent or inter-agent level, mitigating the influence of cumulative drift from previous steps. 
At the intersection of the three, a novel feed-forward model, namely Multi-Agent Global Map Aggregation  (\textbf{\ltog{}}), is responsible for merging submaps both at the intra-agent and inter-agent levels.

We demonstrate through extensive experiments that \method{} achieves higher-quality scene reconstruction, as shown in Figure \ref{fig:teaser}, along with accurate pose estimation at almost 10 FPS. Our contributions can be summarized as follows:

\begin{itemize}
    \item We propose the first multi-agent, feed-forward dense scene reconstruction framework for RGB videos, that incrementally reconstructs 3D point maps in a unified coordinate system without known camera parameters.
    \item We propose a custom Multi-Agent Global Map Aggregation module suitable for sub-map merging at both intra-agent and inter-agent levels, producing global point maps and accurate camera poses.
    \item We lift several existing, feed-forward 3D reconstruction models to the multi-agent setting. Compared with them, \method{} achieves state-of-the-art results in both 3D reconstruction and camera tracking.
\end{itemize}

\section{Related Work}
\label{sec:formatting}

We briefly review the literature concerning 3D reconstruction, classifying existing methods into offline, SLAM, 3R-based systems and multi-agent SLAM.

\textbf{Offline 3D Reconstruction.} Methods belonging to this category assume the availability of a complete set of images beforehand 
\cite{lindenberger2021pixel, liu20233d, schonberger2016structure, snavely2006photo}. If images alone are available, both camera parameters and 3D points can be estimated via Structure from Motion (SfM) \cite{arrigoni2025taxonomy} algorithms. On the contrary, if cameras are known in advance, 3D reconstruction is carried out through Multi-View Stereo (MVS) \cite{wang2024learning}. More modern approaches have been proposed for both settings, ranging from implementing NeRF \cite{mildenhall2020nerf} or 3DGS \cite{kerbl20233d}-like frameworks for 3D surface reconstruction \cite{li2023neuralangelo, chen2024mvsplat, chen2024mvsplat360, guedon2024sugar,chen2024pgsr} in the latter, or training large foundation \cite{wang2025vggt} models in the former. 
Offline processing, however, is often not practical for real systems where agents navigate and interact with the environment.

\textbf{Online SLAM systems.} Traditional SLAM systems like ORB-SLAM \cite{campos2021orb} rely on sparse feature matching, while deep learning approaches such as DROID-SLAM \cite{teed2021droid} use learned features and dense bundle adjustment. However, these methods only produce sparse or semi-dense maps, lacking the capabilities for photorealistic reconstruction.
Neural implicit representations have recently revolutionized dense SLAM systems \cite{tosi2024nerfs,gong2025hs,gong2025dino}. NeRF-based approaches like iMap \cite{sucar2021imap} pioneered the use of MLPs for joint mapping and tracking, while NICE-SLAM \cite{zhu2022nice} introduced hierarchical feature grids to improve scalability. Subsequent works like Vox-Fusion \cite{yang2022vox}, Point-SLAM \cite{sandstrom2023point}, and Co-SLAM \cite{wang2023co} further enhance reconstruction quality through adaptive voxel allocation, neural point clouds, and hybrid coordinate encodings. GO-SLAM \cite{zhang2023go} integrates loop closing and bundle adjustment for global optimization and improved consistency. 
On the other hand, 3DGS has emerged as an alternative explicit representation in this field. GS-SLAM \cite{yan2024gs}, Photo-SLAM \cite{huang2024photo}, SplaTAM \cite{keetha2024splatam}, and MonoGS \cite{matsuki2024gaussian} demonstrate real-time rendering capabilities while maintaining high photorealistic quality. Further works improve upon these methods to enhance tracking accuracy \cite{ha2024rgbd, zhu2025loopsplat, li2025stereo}, mapping efficiency \cite{peng2024rtg}, and leverage multi-modal inputs \cite{sun2024mm3dgs}.
Despite these advances, radiance-field representations remain frame-to-model approaches, mostly operating in the RGB-D setting, that require iterative parameter updates and per-scene optimization, limiting their speed in real-time applications. 

\textbf{Online 3R Models.}
Recent foundation models bypass scene-specific training to offer direct 3D reconstruction \cite{deng2025best3dscenerepresentation}. This began with DUSt3R \cite{wang2024dust3r} predicting point maps from image pairs in an end-to-end manner, which MASt3R \cite{leroy2024grounding} improved via dense feature matching. More recently, VGGT \cite{wang2025vggt} advanced the field by jointly predicting point clouds, poses, and intrinsics from arbitrary image sequences in a single forward pass. Building on these works, novel SLAM systems have emerged. MASt3R-SLAM \cite{murai2025mast3r} leverages MASt3R for real-time monocular SLAM without requiring camera calibration, employing efficient point map matching and Sim(3) optimization. SLAM3R \cite{liu2025slam3r} introduces an end-to-end system with Image-to-Points (I2P) and Local-to-World (L2W) modules for progressive alignment. VGGT-SLAM \cite{maggio2025vggt} adopts the more powerful VGGT transformer, optimizing over the SL(4) manifold to handle projective ambiguities in uncalibrated settings. Spann3R \cite{wang20253d}, instead, uses spatial memory to maintain global consistency during incremental reconstruction, SLAM3R \cite{liu2025slam3r} implements a local to global alignment module, while Ov3R \cite{gong2026ov3r} tackles open-vocabulary 3D semantic reconstruction. 
While these methods are faster than NeRF/3DGS-based SLAM systems and could be optimal candidates to operate in multi-agent settings, existing frameworks operate in single-agent settings and lack robust mechanisms for multi-agent collaboration and global map consistency.

\textbf{Multi-Agent SLAM.} Multi-agent SLAM can be categorized into \textit{centralized} and \textit{distributed} architectures. Traditional centralized systems like CCM-SLAM \cite{schmuck2019ccm} and CVI-SLAM \cite{karrer2018cvi} use a central server for map fusion and global optimization, while distributed approaches such as Swarm-SLAM \cite{lajoie2023swarm} support peer-to-peer communication with robust inter-agent loop closure. Recent multi-agent frameworks like CP-SLAM \cite{hu2023cp} (distributed-to-centralized) and MNE-SLAM \cite{deng2025mne} (peer-to-peer) have integrated neural representations; however, both suffer from the computational overhead inherent to implicit neural methods. MAGiC-SLAM \cite{yugay2025magic} leverages 3DGS for faster rendering and supports multiple agents through centralized coordination, achieving improved tracking via loop closure mechanisms. However, these methods are inherently RGB-D SLAM systems that require known camera parameters, perform slow scene-specific optimization, and focus on view synthesis rather than 3D reconstruction.
In contrast, our \method{} builds upon feed-forward 3R models to achieve multi-agent collaboration without neural representation optimization for the first time. This enables the handling of unknown camera parameters via direct point map prediction at almost 10 FPS, extending applicability to RGB-only online 3D reconstruction.

\begin{figure*}[t]
	\centering
	\includegraphics[width=\linewidth,scale=1.00]{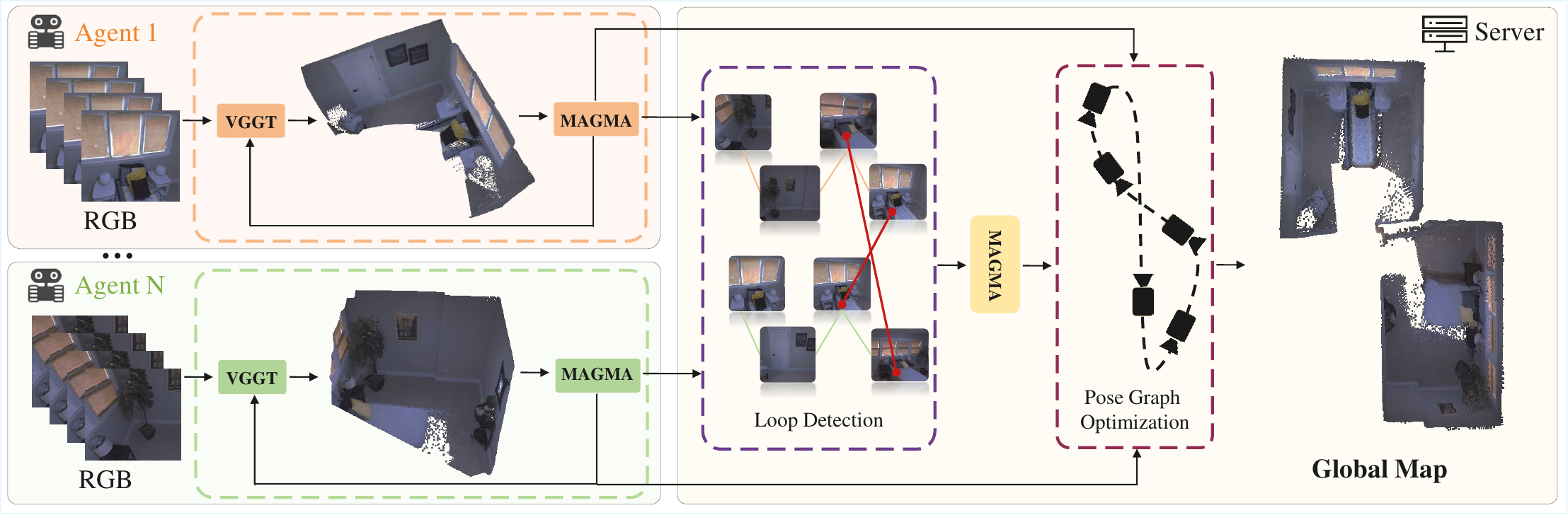}\vspace{-0.3cm}
	\caption{\textbf{Overview of \method{}}. Given RGB sequences from multiple agents, our method simultaneously predicts local point maps and camera poses. Then, the \ltog{} model merges submaps at the intra-agent level. On the server side, when loops are detected between agents, the same \ltog{} module fuses the local maps of multiple agents into a global map. Finally, pose graph optimization further mitigates cumulative drift. }
 \label{fig:overview}
\end{figure*}

\section{Method}

We now introduce \method{}, whose architecture is shown in Figure \ref{fig:overview}.

\subsection{Intra-agent Reconstruction}

We begin by describing how each agent independently processes its RGB sequence to generate local submaps.

\textbf{Incremental Submap Generation}. Following 3R-based systems, each agent in \method{} incrementally reconstructs scenes in its own unified coordinate system by predicting multiple submaps, e.g., $n$. Specifically, for any new set $\bm{I}_i=\left \{ I_{1}^i,...,I_{m}^i \right \} $ of $m$ consecutive RGB images, a submap $\bm{S}_{i}$ is created, while the middle frame $I^i_\frac{m}{2}$ is labeled as a keyframe and added to a set of keyframes $\bm{I}_{key} = \{I^1_\frac{m}{2}, ...,I^n_\frac{m}{2} \}$, used later to support submap merging into a single one.
Purposely, we use a 3R model, VGGT \cite{wang2025vggt}, to encode $\bm{I}_i$ into camera tokens $\bm{t}_i=\left \{ t_{1}^i,...,t_{m}^i \right \}$, and then predict dense depth maps $\bm{D}_i=\left \{ D^i_{1},...,D^i_{m} \right \}$, confidence score maps $\bm{C}_i=\left \{ C^i_{1},...,C^i_{m} \right \}$, and camera intrinsic and extrinsic parameters $\bm{P}_i =\left \{ P^i_{1},...,P^i_{m} \right \}$: 
\begin{equation}
 \bm{t}_{i},\bm{D}_{i},\bm{C}_{i},\bm{P}_{i}  =\Phi _\text{VGGT}(\bm{I}_i) 
\end{equation}
Following \cite{wang2025vggt}, we obtain point maps $\bm{X}_i$ for frames in $\bm{I}_i$ by inverse projecting depth maps $\bm{D}_i$ according to camera parameters $\bm{P}_i$, in the reference system of the first camera. 
We then filter $\bm{X}_i$ according to $\bm{C}_i$, removing any 3D points with confidence lower than $\tau _{conf}$. 

Concurrently, for frames in $\bm{I}_i$ we extract visual tokens $\bm{\upsilon}_{i} = \{\upsilon^i_1, ..., \upsilon^i_m \}  
$ with a ViT encoder $E_\text{ViT}$, as well as spatial tokens
$\bm\vartheta  _{i} = \{\vartheta^i_1, ..., \vartheta^i_m \}$ and global descriptors $\bm\delta_{i} = \{\delta^i_1, ..., \delta^i_m \}$ using DINOv2 SALAD \cite{izquierdo2024optimal}: 

\begin{equation}
    \bm\upsilon_{i} = E_\text{ViT}(\bm{I}_{i}) \quad\quad \bm\delta_{i},  \bm\vartheta  _{i} = E_\text{SALAD}(\bm{I}_{i})
\end{equation}
These tokens are crucial for properly detecting loops and merging submaps into the final map, as we will detail in the remainder. We therefore define single submaps as $\bm{S}_{i} = \left \{ \bm{t}_{i}, \bm{X}_{i}, \bm{P}_{i}, \bm\upsilon_{i}, \bm\delta_{i}, \bm\vartheta_{i} \right \}$ and, concurrently, define $\bm{S}_{key} =\left \{ \bm{t}_{key}, \bm{X}_{key}, \bm{P}_{key}, \bm\upsilon_{key}, \bm\delta_{key}, \bm\vartheta_{key}   \right \}$ by retaining the predictions obtained from frames in $\bm{I}_{key}$. This latter set will be instrumental in performing submap merging and loop closure.

According to this scheme, a single agent extracts $n$ submaps along the scene. However, these have different reference systems -- i.e., each one in the reference system of its first frame $I^i_1$. 
To merge these submaps into a single map in a shared reference system, we introduce our \ltog{} model.

\subsection{\ltog{} Model}\label{sec:magma}

\begin{figure*}[t]
	\centering
	\includegraphics[width=\linewidth,scale=1.00]{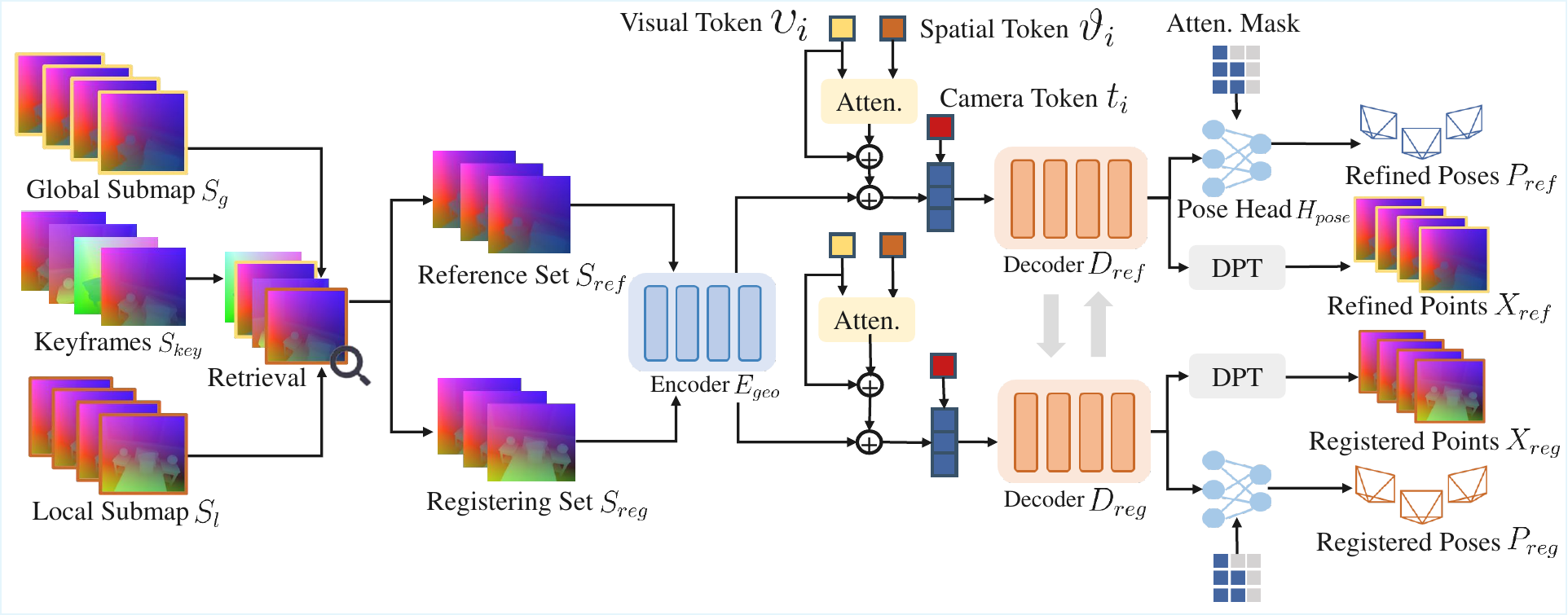}\vspace{-0.3cm}
	\caption{\textbf{Overview of \ltog{} model}. Given a global submap, a keyframe set, and a local submap, the \ltog{} module merges the local submap into a global map with poses in a unified coordinate system. First, we use retrieval to obtain the most relevant frames to form the inputs, reference set and registering set. Next, we encode their point maps and aggregate these geometric features with visual tokens, spatial tokens, and camera tokens. Then, we decode and predict global point maps. 
 }
 \label{fig:magma}
\end{figure*}

Fig. \ref{fig:magma} illustrates \ltog{} (Multi-Agent Global Map Aggregation), which incrementally aligns new submaps into a global coordinate system.  It processes a reference set $\bm{S}_{ref}$ and a registering set $\bm{S}_{reg}$, representing the global map and the newly predicted submap demanding merging, respectively.
These are forwarded to a geometric encoder $E_{geo}$ and two decoders: a registering decoder $D_{reg}$ that predicts point maps and poses aligned to the reference set, and a reference decoder $D_{ref}$ that refines the original reference set.

\textbf{Reference Set Retrieval}. 
Inputs $\bm{S}_{reg}$ and $\bm{S}_{ref}$ to \ltog{} are retrieved respectively from a local submap $\bm{S}_{l}$ demanding merging, and a global set $\left \{ \bm{S}_{g}, \bm{S}_{key} \right \} $. After the single agent has processed images in $\bm{I}_i$, the local submap $\bm{S}_{l}$ to merge is $\bm{S}_{i}$, the global map $\bm{S}_{g}$ contains previously merged submaps $\{\bm{S}_{i-1}, \bm{S}_{i-2}, ...\}$, while the keyframe set $\bm{S}_{key}$ provides historic observations. 
 
To retain only the most relevant views, we select the top-$N$ best-correlated from $\left \{ \bm{S}_{g}, \bm{S}_{key} \right \} $ as the reference set for submap merging. 
Specifically, we measure the visual similarity between descriptors in $\bm\delta_l$ and those in $\bm\delta_g$ and $\bm\delta_{key}$ to obtain correlation scores. Based on these scores, we select the most relevant $N$ views from $\left \{ \bm{S}_{g}, \bm{S}_{key} \right \} $ for $\bm{S}_{ref}$, while we retain $\bm{S}_l$ as the registering set $\bm{S}_{reg}$. 

\textbf{\ltog{} Encoder.} \ltog{} encodes reference and registering point maps $\bm{X}_{ref}, \bm{X}_{reg}$ into geometry tokens $\bm{\gamma}_{ref} = \{ \gamma_1, ..., \gamma_N\}_{ref}$ and $\bm{\gamma}_{reg} = \{ \gamma_1, ..., \gamma_m\}_{reg}$
\begin{equation}
    \bm{\gamma}_{ref} = E_{geo}(\bm{X}_{ref}), \quad \bm{\gamma}_{reg} = E_{geo}(\bm{X}_{reg})
\end{equation}
$E_{geo}$ includes a 2D conv layer with kernel size 16 and stride 16, similar to the ViT patch embedding process, making the geometry tokens $\gamma_{i}$ spatially aligned with visual tokens $\bm{\upsilon}_i$ to capture structural geometric primitives. However, geometry features alone may be insufficient for merging due to repetitive structures and lack of texture in point maps. 

To address this issue, we introduce appearance tokens $\bm{\alpha}_{ref}  = \{\alpha_{1}, ..., \alpha_{N} \}_{ref}$ for the reference set to enhance the geometry tokens and establish correspondences across different coordinate systems. These are derived from the previously introduced visual and spatial tokens through an attention module:
\begin{equation}
    \bm{\alpha}_{ref} = \bm\upsilon_{ref} + softmax(\frac{\bm\upsilon_{ref}\bm\vartheta _{ref}^{T}}{\sqrt{d} } )\bm\vartheta  _{ref}
\end{equation}
where $d$ denotes the feature dimension. We apply the same process to obtain $\bm{\alpha}_{reg}$.
The appearance tokens establish correspondences across different coordinate systems and mitigate the ambiguity inherent in geometry tokens alone.
Geometry and appearance tokens are then combined and concatenated with camera tokens into integrated features $\bm{F}_{ref} = \{\mathcal{F}_1, ..., \mathcal{F}_{N} \}_{ref}$:
\begin{equation}
    \bm{F}_{ref}=[\bm{t}_{ref}; \bm{\alpha}_{ref}+\bm{\gamma}_{ref}]
\end{equation}
We apply the same process to obtain $\bm{F}_{reg}$.

\textbf{\ltog{} Decoders}. Registering and reference decoders $D_{reg},D_{ref}$ process $\bm{F}_{ref}, \bm{F}_{reg}$, respectively. Each decoder block contains self-attention and multi-view cross-attention layers, followed by an MLP. The registering decoder first applies self-attention with $\bm{F}_{reg}$, then performs cross-attention with $\bm{F}_{reg}$ as queries and $\bm{F}_{ref}$ as keys and values. 
Finally, after max-pooling, we obtain registering tokens $\bm{G}_{reg} = \{\mathcal{G}_1, ..., \mathcal{G}_m \}_{reg}$ and reference tokens $\bm{G}_{ref} = \{\mathcal{G}_1, ..., \mathcal{G}_N \}_{ref}$:
\begin{equation}
    \bm{G}_{reg}=D_{reg}(\bm{F}_{reg}, \bm{F}_{ref}), \quad
    \bm{G}_{ref}=D_{ref}(\bm{F}_{ref}, \bm{F}_{reg})
\end{equation}

\textbf{\ltog{} Heads}. Finally, DPT heads \cite{ranftl2021vision} $H_{pts}$ regress the refined point map ${\bm{X}'}_{ref}$, the registered point map ${\bm{X}'}_{reg}$, and confidence maps $\bm{C}_{ref}, \bm{C}_{reg}$ 
\begin{equation}
    \bm{X'}_{ref}, \bm{C}_{ref} = H_{pts}(\bm{G}_{ref}), \quad \bm{X'}_{reg}, \bm{C}_{reg} = H_{pts}(\bm{G}_{reg})
\end{equation}
By retrieving pose embeddings $\bm{t}_{ref}, \bm{t}_{reg}$ from $\bm{G}_{ref}, \bm{G}_{reg}$, 
we use a pose head $H_{pose}$, consisting of four self-attention layers followed by a linear layer, to predict the refined camera parameters $\bm{P}_{ref}$ and the registered camera parameters $\bm{P}_{reg}$, with the latter transformed into the global coordinate system. Specifically, we concatenate the pose embeddings as $(\bm{t}_{ref}, \bm{t}_{reg})$. Then, we use an attention mask in the pose head to guide the information exchange: $\bm{t}_{ref}$ attends to all its tokens, while $\bm{t}_{reg}$ attends only to tokens in $\bm{t}_{ref}$.
 \begin{equation}
    \bm{P}_{ref}, \bm{P}_{reg} = H_{pose}(\bm{t}_{ref},\bm{t}_{reg}) 
\end{equation}
\normalsize 
The global map is then updated in its point maps and poses accordingly.

\textbf{Training Loss}. We train \ltog{} end-to-end by supervising submap fusion. Following DUSt3R \cite{wang2024dust3r}, we compute a confidence-aware registration loss between predicted point maps $\bm{X}_g$ in the global coordinate system and the ground truth point map $\hat{\bm{X}}_{g}$, normalized by the average Euclidean distance of valid points:
\begin{equation}
\mathcal{L}_{reg}= 
M \cdot \Big( \bm{C}_{g}\cdot \left \| \bm{X}_{g} -\hat{\bm{X}}_{g} \right \| -\beta \log\bm{C}_{g} \Big)
\end{equation}
where $M$ is the mask of valid points with ground truth, $z$ and $\hat{z}$ are scale factors, and $\beta$ is a hyperparameter for regularization. 
We also enforce a camera loss between camera parameters $\bm{P}_{g}$ and ground truth ones $\hat{\bm{P}_{g}}$ as
\begin{equation}
    \mathcal{L}_{pose}= \left \| \bm{P}_{g}-\hat{\bm{P}_{g}} \right \|
\end{equation}
where $\hat{\bm{P}}_{g}=[q, b, f]$ contains a quaternion $q\in \mathbb{R}^{4}$, the translation vector $b\in \mathbb{R}^{3}$, and the field of view $f\in \mathbb{R}^{2}$, assuming the principal point at the image center.

Finally, to ensure that the registered submap is consistently aligned with the global scene, we add a geometry consistency term. 
By randomly selecting two views $j,k$ in the predicted submaps, we use ground truth depth maps $\hat{{D}}_j,\hat{{D}}_k$ and intrinsics $\hat{{K}}_j,\hat{{K}}_k$, together with the  predicted camera poses ${T}_j,{T}_k$ derived from $\bm{P}_g$, to reproject pixels from view $j$ to view $k$. We then measure the distance between projected point maps ${\tilde{X}}_{j}$ and ${\tilde{X}}_{k}$ in the global coordinate system as:
\begin{equation}
\mathcal{L}_{geo}= 
M \left \| {\tilde{X}}_{j} - {\tilde{X}}_{k}    \right \| _{2}
\end{equation}
where $M$ is the mask of valid pixels.
The total training loss $\mathcal{L}$ is then defined as:
\begin{equation}
\mathcal{L}=\lambda_{reg} \mathcal{L}_{reg} + \lambda_{pose} \mathcal{L}_{pose} + \lambda_{geo} \mathcal{L}_{geo}
\end{equation}

\subsection{Inter-agent Reconstruction}

We now present how the \ltog{} model combines the efforts of the single agents into a global 3D reconstruction. 

\textbf{Loop Detection}. 
To achieve global and consistent reconstruction across multiple agents, we need to detect regions where the trajectories of the different agents overlap. We set the first camera of the first agent $A_1$ as the global coordinate system. Then, when a new submap is created by another agent, e.g., $A_2$, we look for loops between this submap and keyframes $\bm{S}_{key}^{A_1}$ from $A_1$, 
by computing correlations between ${\bm{\delta}^{{A_2}}_i}$ and ${\bm{\delta}^{{A_1}}_{key}}$.
When the maximum correlation score exceeds a threshold $\tau _{loop}$, a loop between the agents is detected.

\textbf{Inter-agent Submap Fusion}. Once a loop is detected, we take the submap from the first agent $A_1$ containing the keyframe with the maximum correlation score as the global submap $\bm{S}^{A_1}_{g}$ and forward it to \ltog{}, together with local submap $\bm{S}^{A_2}_{l}$ (the latest submap reconstructed by the other agent $A_2$) and the keyframe set $\bm{S}^{A_2}_{key}$. 
We retrieve the top-$N$ most-correlated views from the $\left \{ \bm{S}^{A_2}_{l}, \bm{S}^{A_2}_{key} \right \}$ as the registering set $\bm{S}_{reg}$, where $\bm{S}^{A_2}_{key}$ serves as historic observations providing global clues. Concurrently, we retain $\bm{S}^{A_1}_{g}$ as the reference set $\bm{S}_{ref}$. Then, \ltog{} predicts the registered point map and poses for agent $A_2$, as described in Sec. \ref{sec:magma}, which are further used to compute the relative transformation between local and global coordinates and update the global map. 

\subsection{Backend Pose Graph Optimization}

We further refine predicted poses through classical optimization techniques \cite{yugay2025magic}.

\textbf{Graph Construction}. To mitigate cumulative errors during the incremental process, we perform pose graph optimization (PGO) after submap merging to reduce drift and enforce global consistency. Each submap created by an agent corresponds to a node $\bm{n}_{i} \in SE(3)$ in the graph. The edges $e_{ij}$ between the neighboring nodes represent the relative transformations computed by the proposed \ltog{} model during the intra-agent stage. 

\textbf{Loop Closure}. We compute the visual similarity between the extracted global descriptors $\bm{\delta}_{i}$ of the keyframes in each agent to retrieve potential loop candidates for a new submap. If a correlation score higher than $\tau _{loop}$ is found, a loop is detected. The loop edge is estimated by \ltog{} during the inter-agent stage. 

\textbf{Optimization}. We obtain the optimized camera poses by leveraging the Levenberg-Marquardt algorithm to minimize the objective function:
\begin{equation}
    \min\sum_{i,j \in \varepsilon } \left \| log (e_{ij}^{-1}(n_{i}^{-1}n_{j}))\right \| ^{2}_{\Omega _{ij}}
\end{equation}
where $\varepsilon$ denotes the index of nodes. $n_{i}$ and $n_{j}$ are node poses, and $\log$ is the logarithmic map from $SE(3)$ to $\mathfrak{se}(3)$. $\Omega _{ij}$ is the information matrix reflecting the uncertainty over edges -- the higher the score, the higher the weight. 

\textbf{Pose Update Integration}. After performing pose graph optimization, we obtain the pose corrections $\bm{n'}_{i}$ for each submap of each agent. All camera poses $\bm{T}_i=\left \{ T^i_{1},...,T^i_{m} \right \} $ in each submap are updated as:
\begin{equation}
\bm{T}_{i} = \bm{n'}_{i} \bm{T}_{i}
\end{equation}
\section{Experiments}

We now present a thorough evaluation of \method{}, including implementation details, ablation studies, and comparisons with state-of-the-art methods. 

\textbf{Dataset}. We train our Multi-Agent Global Map Aggregation (\ltog{}) module on a mixture of several datasets: ScanNet \cite{dai2017scannet}, ScanNet++ \cite{yeshwanth2023scannet++}, and Aria Synthetic Environment \cite{avetisyan2024scenescript}. ScanNet and ScanNet++ provide real-world, diverse indoor environments, ranging from small rooms to large-scale indoor scenarios. Aria Synthetic Environment offers photorealistic multi-room scenes. 
To effectively learn how to merge point maps both at the intra-agent and inter-agent levels, \ltog{} is trained by partitioning the diverse scenes in the training set to simulate a multi-agent system -- since none of the training datasets support this setting natively. Specifically, two trajectories are created starting from the first and last frames of a scene, sampling one frame every 2 for each trajectory. 

For evaluation, we perform 3D reconstruction and camera tracking on the ReplicaMultiagent \cite{hu2023cp} and the AriaMultiagent \cite{yugay2025magic} datasets. The former offers synthetic single and multiple rooms where 2 agents navigate independently and is specifically tailored for 3D reconstruction, while the latter provides real-world indoor environments explored by 3 agents, with ground truth camera information and depth available for evaluation, yet no ground truth reconstructions. 

\textbf{Implementation Details.} We implement \method{} in PyTorch. Specifically, we train \ltog{} for 200 epochs with batch size 5 on 4 A100 GPUs and learning rate 1.5e$^{-5}$. We set the number of input images processed by VGGT $m=10$ with stride $s=5$, the confidence threshold $\tau _{conf}=0.25$, the correlation threshold $\tau _{loop}=0.6$, top-correlated samples $N=10$, the regularization term $\beta=1$, and the loss weights $\lambda_{reg}=1, \lambda_{pose}=1, \lambda_{geo}=0.8$. Following the literature on multi-agent SLAM \cite{deng2025mne,yugay2025magic}, we run our evaluation in a multi-agent setting, while also reporting the results achieved by the single agents running independently.

\textbf{Baselines.} For single-agent evaluation, we compare our \method{} against state-of-the-art feed-forward frameworks: DUSt3R \cite{wang2024dust3r} and MASt3R \cite{leroy2024grounding} processing two views at a time sequentially, SLAM3R \cite{liu2025slam3r}, MASt3R-SLAM \cite{murai2025mast3r}, and VGGT-SLAM \cite{maggio2025vggt} as incremental multi-view 3D reconstruction methods. For the multi-agent setting, existing neural SLAM systems \cite{hu2023cp,deng2025mne,yugay2025magic} require RGB-D input; therefore, we report them as references rather than fair baselines when evaluating tracking accuracy. Instead, we adapt \cite{wang2024dust3r, leroy2024grounding, liu2025slam3r, maggio2025vggt} for multi-agent operation by independently processing each agent's video and merging the resulting point maps using RANSAC + ICP alignment (RICP), providing comparable feed-forward baselines, and we compare against a concurrent work lifting MASt3R-SLAM \cite{murai2025mast3r} to multi-agent setting through global graph optimization -- referred to as MA-MASt3R-SLAM \cite{zhou2025multi}. 

\textbf{Metrics}. For quantitative evaluation, we follow \cite{liu2025slam3r} and build ground truth point clouds by projecting pixels to 3D points using ground truth depth and camera parameters. Reconstruction quality is evaluated through accuracy and completeness, tracking accuracy using Absolute Trajectory Error (ATE) RMSE. 

\subsection{Ablation Studies}
We start our evaluation by running ablation experiments on ReplicaMultiagent. 

\begin{wraptable}{l}{0.45\textwidth}\vspace{-1cm}
\caption{\textbf{Ablation study -- Impact of inter-agent merging strategies.} 
Results in the multi-agent setting.}
\resizebox{0.45\columnwidth}{!}{
\begin{tabular}{lcrrr}
\toprule
Methods       & Train Setting & Acc.          & Comp.         & RMSE          \\ \midrule
(A) w/ RICP          & $\times$             & \trd 8.89          & \trd 6.47          & \trd 7.66          \\
(B) w/ SL(4)         & $\times$             & 13.66         & 11.67         & 18.94         \\
(C) w/ L2W           & Multi Agent            & 9.71          & 7.16          & 9.19          \\
\midrule
(D) w/ \ltog{}          & Single Agent            & \snd 6.89          & \snd 4.93          & \snd 5.35          \\
(E) \textbf{\method{}} & Multi Agent            & \fst \textbf{5.37} & \fst\textbf{3.36} & \fst\textbf{3.87} \\ 
\bottomrule \\ 
\toprule
(F) w/o L$_{geo}$ & Multi Agent & 5.88 & 3.60 & 3.97 \\
(G) w/o Spatial Tokens & Multi Agent & 6.04 & 3.71 & 4.19 \\
\bottomrule
\end{tabular}
}\vspace{-0.5cm}
\label{tab:ablation_merging}
\end{wraptable}

\textbf{Impact of \ltog{}}. Tab. \ref{tab:ablation_merging} demonstrates the effectiveness of \ltog{} at combining the submaps from different agents, by comparing it with different merging strategies on the ReplicaMultiagent dataset. In this experiment, we report both reconstruction and camera tracking metrics in the multi-agent setting. We select three typical fusion strategies as baselines, including (A) RANSAC+ICP (RICP) used by MAGiC-SLAM \cite{yugay2025magic}, (B) optimization over SL(4) manifold used by VGGT-SLAM \cite{maggio2025vggt}, and (C) the L2W module proposed in SLAM3R \cite{liu2025slam3r}. 
Finally, (D) we also evaluate the effectiveness of \ltog{} in the multi-agent setting, yet trained in a single-agent regime only. All variants use VGGT as the single-agent backbone.
We can notice how \ltog{} consistently outperforms existing alternatives, even when not trained to deal with multiple agents (D) -- while a proper training under multi-agent assumptions unleashes its full potential (E). 
We observe that both RICP (A) and SL(4) (B) heavily rely on the similarity of two fused submaps: the less similar the two subsets, the poorer the alignment obtained when merging. This makes them well-suited for incremental intra-agent alignment, yet ineffective for operating at the inter-agent level. 
In contrast, our \ltog{} module learns corresponding matches using geometry features across the subsets, remaining effective even with lower overlaps. The L2W branch from SLAM3R \cite{liu2025slam3r} also underperforms at this task, due to the poorer geometry and visual information it processes.
On a side note, RICP achieves the best results among the other methods, motivating our choice of using it to build the multi-agent feed-forward baselines for the rest of our experiments.
Finally, we report the results achieved by ablated versions of MAGMA in (F) and (G), respectively, by removing L$_{geo}$ loss and the spatial tokens, confirming that both play an important role.

\begin{wraptable}{r}{0.45\textwidth}\vspace{-1cm}
\caption{\textbf{Ablation study -- Impact of backbones and PGO}. Results in the multi-agent setting.}
\resizebox{0.45\columnwidth}{!}{%
\setlength{\tabcolsep}{10pt}
\begin{tabular}{lrrr}
\toprule
\multicolumn{1}{c}{Methods}       & Acc.          & Comp.         & RMSE          \\ \midrule
(E) \textbf{\method} & \fst \textbf{5.37} & \fst \textbf{3.36} & \fst \textbf{3.87} \\
(H) w/ SAIL-Recon \cite{deng2025sail}                 & \snd 5.97          &  \snd 3.88         & \snd 4.06          \\
(I) w/o PGO                       & \trd 6.18          &  \trd 4.11         & \trd 5.65          \\ \bottomrule
\end{tabular}}
\vspace{-0.6cm}
\label{tab:ablation_pgo}
\end{wraptable}
\textbf{Effects of Backbone}. We compare using VGGT against an alternative backbone, the recent state-of-the-art model SAIL-Recon \cite{deng2025sail}. 
This experiment is reported in Tab. \ref{tab:ablation_pgo} row (H). 
The results shown by our original framework (E) demonstrate that selecting VGGT yields better results.

\textbf{Effects of PGO}. Finally, we measure the impact of Pose Graph Optimization (PGO) on our SLAM system. 
As shown in Tab. \ref{tab:ablation_pgo} row (I), removing this component yields significant drops in accuracy, making it more impactful than backbone selection. However, by comparing (I) with entries (A), (B), and (C) in Tab. \ref{tab:ablation_merging}, we can appreciate how \ltog{} still outperforms existing merging strategies even without PGO.

\begin{table*}[t]
\centering
\caption{\textbf{Reconstruction results on ReplicaMultiagent}. 
$^\amalg$ means using RICP at the inter-agent level. The best results are shown as \colorbox[HTML]{BDE6CD}{\textbf{first}}, \colorbox[HTML]{E2EEBC}{second}, and \colorbox[HTML]{FFF3BB}{third}. }
\label{tab:replica_recon}\vspace{-0.3cm}
\renewcommand{\tabcolsep}{8pt}
\resizebox{\textwidth}{!}{%
\begin{tabular}{llcrrrrrrrrrr}
\toprule
\multirow{2}{*}{Methods} &
  \multirow{2}{*}{Setting} &
  \multicolumn{2}{c}{Apart 0} &
  \multicolumn{2}{c}{Apart 1} &
  \multicolumn{2}{c}{Apart 2} &
  \multicolumn{2}{c}{Office 0} &
  \multicolumn{2}{c}{Avg.} \\ 
  \cmidrule{3-12} 
 &
   &
  Acc. &
  Comp. &
  Acc. &
  Comp. &
  Acc. &
  Comp. &
  Acc. &
  Comp. &
  Acc. &
  Comp. \\ \midrule
DUSt3R\cite{wang2024dust3r} &
  \multirow{5}{*}{Agent 1} &
  6.95 &
  3.85 &
  15.78 &
  12.29 &
  10.15 &
  9.17 &
  12.63 &
  12.43 &
  11.38 & 9.44 \\
MASt3R\cite{leroy2024grounding} &
   &
  \trd 5.01 &
  \trd 3.00 &
  \trd 9.51 &
  \trd 3.98 &
  \trd 7.32 &
  \trd 3.91 &
  \trd 6.14 &
  \trd 5.30 & 
  \snd 7.00 & \snd 4.05 \\
SLAM3R\cite{liu2025slam3r} &
   &
  5.17 &
  4.03 &
 \snd 9.18 &
  6.91 &
  \snd 4.97 &
  4.94 &
  10.18 &
  7.10 & 
 \trd 7.38 & 5.75 \\
MASt3R-SLAM\cite{murai2025mast3r} &
 &
  10.91 &
  6.35 &
  10.38 &
 \snd 3.24 &
  9.68 &
 \fst 3.28 &
  7.58 &
  6.91 &
  9.64 &  \trd 4.94 \\
VGGT-SLAM\cite{maggio2025vggt} &
   &
  \snd 4.59 &
  \snd 2.89 &
  15.64 &
   9.66 &
  9.87 &
   4.15 &
  \snd 6.06 &
  \snd 4.80 & 
   9.04 &  5.38 \\
\textbf{\method{} (ours)} &
   &
  \fst\textbf{4.07} &
  \fst\textbf{2.56} &
  \fst\textbf{5.57} &
  \fst\textbf{3.05} &
  \fst\textbf{3.96} &
  \snd{3.35} &
  \fst\textbf{5.01} &
  \fst\textbf{2.95} & 
  \fst\textbf{4.65} & \fst\textbf{2.98} \\ \midrule
DUSt3R\cite{wang2024dust3r} &
  \multirow{5}{*}{Agent 2} &
  6.87 &
  4.69 &
  \snd 14.14 &
  \snd 11.82 &
  \trd 7.88 &
  \fst 5.02 &
  \trd 6.98 &
  \trd 6.26 &
 \trd 8.97 & \trd 6.95 \\
MASt3R\cite{leroy2024grounding} &
   &
  \trd 4.95 &
  \trd 2.48 &
  15.78 &
  12.87 &
  10.87 &
  12.39 &
  7.66 &
  7.68 &
  9.82 & 8.86 \\
SLAM3R\cite{liu2025slam3r} &
   &
  7.24 &
  7.68 &
  15.18 &
  14.80 &
  8.73 &
  8.73 &
  \snd 5.98 &
  \fst 4.36 &
   9.28 &  8.89 \\
MASt3R-SLAM\cite{murai2025mast3r} &
 &
  13.09 &
  4.66 &
  21.57 &
  23.93 &
  13.19 &
  14.91 &
  8.06 &
  8.31 &
  13.98 & 12.95 \\
VGGT-SLAM\cite{maggio2025vggt} &
   &
  \snd 4.29 &
  \snd 2.37 &
  \trd 14.41 &
  \trd 11.95 &
  \snd 6.99 &
  \trd 6.15 &
  7.45 &
  6.78 &
  \snd 8.29 & \snd 6.81 \\
\textbf{\method{} (ours)} &
   &
  \fst\textbf{3.43} &
  \fst\textbf{1.79} &
  \fst\textbf{9.26} &
  \fst\textbf{8.94} &
  \fst\textbf{6.89} &
  \snd{5.72} &
  \fst\textbf{5.26} &
  \snd{4.57} &
  \fst\textbf{6.21} & \fst\textbf{5.26} \\ 
   \midrule\midrule
DUSt3R\cite{wang2024dust3r} $^\amalg$ &
  \multirow{5}{*}{Multi-Agent} &
  11.79 &
  8.72 &
  \trd 15.21 &
   11.71 &
  \trd 11.24 &
  10.24 &
  11.80 &
  9.03 &
  12.51 & 9.92 \\
MASt3R\cite{leroy2024grounding} $^\amalg$ &
   &
  \snd 6.29 &
  \snd 4.83 &
  \snd 12.97 &
  \snd 9.31 &
  12.03 &
  10.48 &
  \trd 9.18 &
  8.01 &
  \snd 10.12 & \snd 8.16 \\
SLAM3R\cite{liu2025slam3r} $^\amalg$ &
   &
  18.09 &
  10.79 &
   16.57 &
  15.25 &
  14.38 &
  \snd 6.44 &
  \snd 8.11 &
  \snd 3.75 &
  14.29 &  9.06 \\
MASt3R-SLAM\cite{murai2025mast3r} $^\amalg$ &
 &
  15.76 &
  6.52 &
  16.26 &
 \trd 10.81 &
  12.69 &
  10.97 &
  10.59 &
  9.16 &
  13.83 & 9.37 \\
VGGT-SLAM\cite{maggio2025vggt} $^\amalg$ &
   &
  \trd 8.60 &
  \trd 5.92 &
  15.97 &
  13.00 &
  \snd 10.23 &
  \trd 6.86 &
  10.65 &
  \trd 7.97 &
  \trd 11.36 & \trd 8.44 \\
\textbf{\method{} (ours)} &
   &
  \fst\textbf{4.20} &
  \fst\textbf{2.57} &
  \fst\textbf{8.32} &
  \fst\textbf{4.26} &
  \fst\textbf{5.79} &
  \fst\textbf{4.61} &
  \fst\textbf{3.18} &
  \fst\textbf{2.01} &
  \fst\textbf{5.37} & \fst\textbf{3.36} \\ \bottomrule
\end{tabular}%
}
\end{table*}

\begin{figure*}[t]
    \centering
    \begin{overpic}[width=0.9\linewidth]{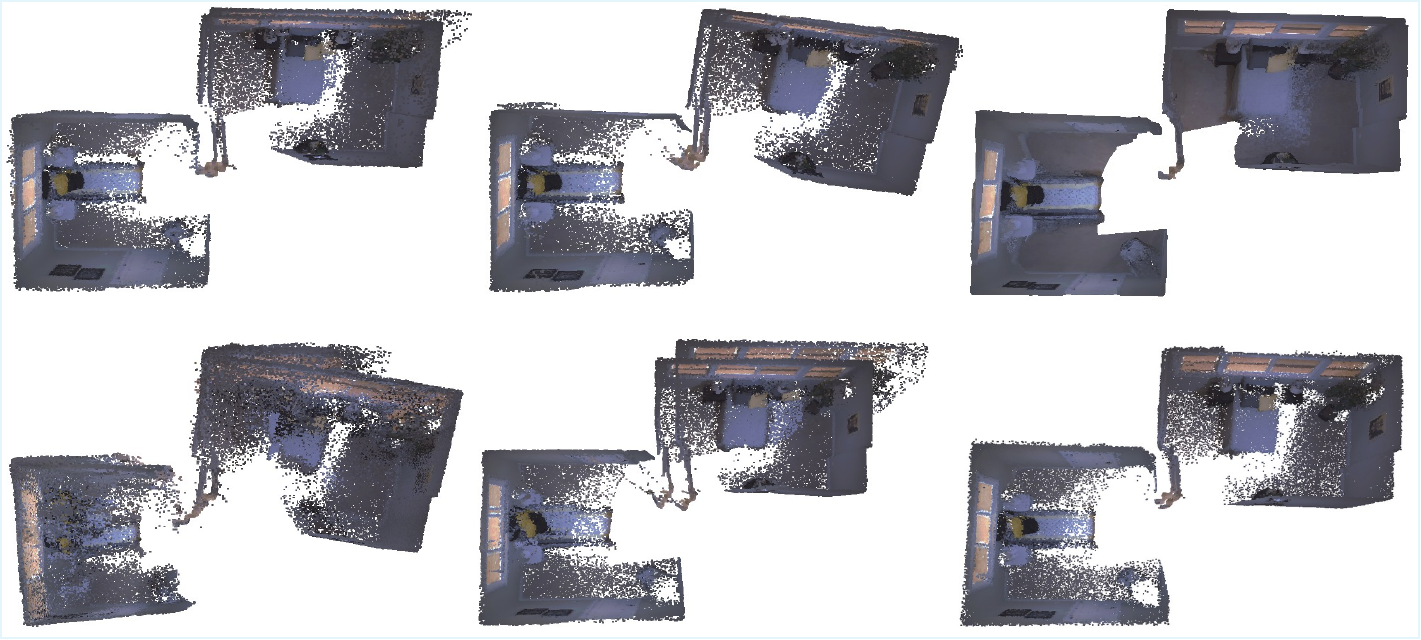}
    \put (80,22.2) {\tiny Ground Truth}
    \put (43,22.2) {\tiny MASt3R \cite{leroy2024grounding} + RICP}
    \put (8,22.2) {\tiny VGGT-SLAM \cite{maggio2025vggt} + RICP}
    \put (74,-1.7) {\tiny \textcolor{purple}{\textbf{\method{} (Ours) }}}
    \put (43,-1.7) {\tiny DUSt3R \cite{wang2024dust3r} + RICP}
    \put (8,-1.7) {\tiny SLAM3R \cite{liu2025slam3r} + RICP}
    \end{overpic}
    \caption{\textbf{Qualitative results -- Dense point maps on ReplicaMultiagent.} We present the reconstruction results with the multi-agent setting on the \emph{Apart 0} sequence. Our method yields a more accurate and consistent global point map.}
    \label{fig:vis}
\end{figure*}
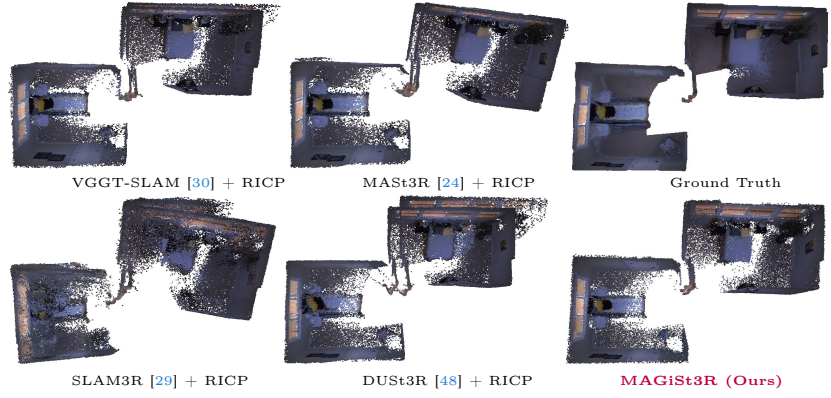

\begin{table}[t]
\caption{\textbf{Tracking results on ReplicaMultiagent (single-agent).}}
\vspace{-0.3cm}
\label{tab:replica_tracking_single}
\renewcommand{\tabcolsep}{2.5pt}
\resizebox{\columnwidth}{!}{%
\begin{tabular}{lcccccccccc}
\toprule
\multirow{2}{*}{Method} & \multicolumn{5}{c}{Agent 1} & \multicolumn{5}{c}{Agent 2} \\ 
\cmidrule(r){2-6} \cmidrule(l){7-11} 
 & Apart 0 & Apart 1 & Apart 2 & Office 0 & {Avg.} & Apart 0 & Apart 1 & Apart 2 & Office 0 & {Avg.} \\ \midrule
DUSt3R \cite{wang2024dust3r} & 6.76 & 8.87 & 7.30 & 21.04 & 10.99 & 8.18 & 9.93 & 10.97 & 4.31 & 8.35 \\
MASt3R \cite{leroy2024grounding} & \fst\textbf{2.52} & \snd 4.46 & \fst\textbf{3.62} & \trd 9.56 & \snd 5.04 & \trd 3.58 & \fst\textbf{3.10} & \snd 3.45 & \fst\textbf{2.08} & \snd 3.05 \\
SLAM3R \cite{liu2025slam3r} & 7.25 & 30.62 & 8.77 & 18.43 & 16.27 & 14.12 & 27.94 & 11.35 & 18.65 & 18.02 \\
MASt3R-SLAM \cite{murai2025mast3r} & 4.19 & \fst\textbf{3.79} &  6.04 & 10.57 & \trd 6.15 & 7.90 & 9.29 & \trd 5.02 & 5.40 & 6.90 \\
VGGT-SLAM \cite{maggio2025vggt} & \trd 2.96 & 8.90 & \trd 5.77 & \snd 9.07 & 6.68 & \snd 2.64 & \trd 7.86 & 5.89 & \trd 3.63 & \trd 5.01 \\
\bf \method{} (ours) & \snd 2.54 & \trd 6.99 & \snd 3.88 & \fst\textbf{2.69} & \fst\textbf{4.03} & \fst\textbf{1.87} & \snd 4.06 & \fst\textbf{3.23} & \snd 2.48 & \fst\textbf{2.91} \\ \bottomrule
\end{tabular}%
}
\end{table}

\subsection{Comparison with the State-of-the-Art}

We now compare \method{} with existing feed-forward models. 

\textbf{3D Reconstruction on ReplicaMultiagent.} In Tab. \ref{tab:replica_recon}, we compare the 3D reconstruction quality of our method against the state-of-the-art feed-forward approaches on the ReplicaMultiagent dataset, containing both single-room and multi-room environments. 
On top, we report the reconstruction accuracy and completion achieved by single agents independently -- i.e., without any inter-agent communication -- followed at the bottom by the results achieved under the multi-agent setting. It is worth highlighting that results achieved by single agents alone and by the multi-agent system as a whole are not directly comparable, as they cover different portions of the entire scene.
By focusing on single agents, \method{} already outperforms any feed-forward model in both Agent 1 and Agent 2 tracks, with few exceptions -- completeness on \emph{Apart 2} and \emph{Office 0}, supporting the superior capability of \ltog{} to aggregate submaps at the intra-agent level. 
Furthermore, when moving to the multi-agent setting, our framework shines and largely outperforms all proposed baselines, confirming the effectiveness of our \ltog{} model even at the inter-agent level, and therefore establishing it as a key component for both single and multiple agent systems.
This fact is confirmed qualitatively by Fig.  \ref{fig:vis}: reconstructions by \method{} expose significantly fewer artefacts with respect to the other methods that, on the contrary, struggle at properly aligning the two rooms explored in the \textit{Apart 0} sequence, often reconstructing duplicated portions of the scene or with some parts being significantly drifted with respect to the global point map.

\begin{wraptable}{r}{0.5\textwidth}\vspace{-1cm}
\caption{\textbf{Tracking results on ReplicaMultiagent (multi-agent).} 
*means average excluding \textit{Apart 2} to compare with \cite{zhou2025multi}.}
\label{tab:replica_tracking_multi}
\resizebox{0.5\columnwidth}{!}{%
\begin{tabular}{lccccc}
\toprule
\multirow{2}{*}{Methods} & \multicolumn{5}{c}{Multi-Agent}                                               \\ \cline{2-6} 
                         & Apart 0       & Apart 1       & Apart 2       & Office 0      & Avg.          \\ \midrule
\multicolumn{6}{c}{RGB-D (Calibrated)}                                                                    \\ \midrule
Swarm-SLAM \cite{lajoie2023swarm}              & 1.80          & 5.56           & 5.61          & 1.42          & 3.60          \\
CP-SLAM \cite{hu2023cp}                 & 0.95          & 1.42          & 1.91          & 0.65         & 1.23          \\
MAGiC-SLAM \cite{yugay2025magic}             & 0.16          & 0.26          & 0.32          & 0.27          & 0.25          \\ \midrule
\multicolumn{6}{c}{RGB (Uncalibrated)} \\ \midrule
DUSt3R \cite{wang2024dust3r} $^\amalg$                  & 8.35          & 12.27         & 11.43         & 14.28         & 11.58         \\
MASt3R \cite{leroy2024grounding} $^\amalg$                   & \trd 4.02          & \trd 6.58          & \snd 5.66          & 7.69          & \snd 5.99          \\
SLAM3R \cite{liu2025slam3r} $^\amalg$                   & 11.72         & 31.92         & 11.74         & 20.40         & 18.94         \\
MASt3R-SLAM \cite{murai2025mast3r} $^\amalg$              & 6.96          & 8.85          & 7.46          & 9.44          & 8.17          \\
VGGT-SLAM \cite{maggio2025vggt} $^\amalg$                & \snd 3.42          & 10.34         & \trd 7.39          & \trd 7.46          & \trd 7.15          \\
MA-MASt3R-SLAM \cite{zhou2025multi}         & 5.24          & \bf \fst 5.48          & -             & \bf \fst 2.57          & - /4.43*            \\
\bf \method{} (ours)   & \fst \textbf{2.75} & \snd {5.81} & \fst {3.84} & \snd {3.09} & \fst \textbf{3.87}/3.88* \\ 
\bottomrule
\end{tabular}%
}\vspace{-0.3cm}
\end{wraptable}

\textbf{Tracking on ReplicaMultiagent.} Tab. \ref{tab:replica_tracking_single} reports tracking accuracy on ReplicaMultiagent, by the same methods when deployed in single-agent settings. Although \method{} achieves mixed results on individual scenes, our method still achieves the average best performance with both agents. 
Furthermore, Tab. \ref{tab:replica_tracking_multi} extends to the multi-agent setting, also reporting existing RGB-D SLAM systems designed for this specific setup \cite{lajoie2023swarm,hu2023cp,yugay2025magic} as upper bounds. In this case, \method{} consistently outperforms all the other multi-agent feed-forward baselines, confirming that \ltog{} effectively improves inter-agent aggregation. On average, \method{} also outperforms the concurrent MA-MASt3R-SLAM \cite{zhou2025multi}, getting very close to RGB-D systems.

\textbf{AriaMultiagent}. 
We extend our evaluation to AriaMultiagent, a real-world egocentric dataset supporting experiments for up to 3 agents. 
In Tab. \ref{tab:arial_tracking_single}, we report tracking accuracy achieved by feed-forward methods running independently in the single-agent settings. Unlike the ReplicaMultiagent experiments, in this case \method{} consistently outperforms other feed-forward methods on both \textit{Room 0} and \textit{Room 1} scenes, along any of the single agent trajectories.

\begin{table}[t]
\caption{\textbf{Tracking results on AriaMultiagent (single-agent). } 
}\vspace{-0.3cm}
\label{tab:arial_tracking_single}
\resizebox{\columnwidth}{!}{%
\renewcommand{\tabcolsep}{5pt}
\begin{tabular}{lccccccccc}
\toprule
\multirow{2}{*}{Methods} & \multicolumn{3}{c}{Agent 1} & \multicolumn{3}{c}{Agent 2} & \multicolumn{3}{c}{Agent 3} \\ 
\cmidrule(r){2-4} \cmidrule(lr){5-7} \cmidrule(l){8-10} 
 & Room 0 & Room 1 & {Avg.} & Room 0 & Room 1 & {Avg.} & Room 0 & Room 1 & {Avg.} \\ \midrule
DUSt3R \cite{wang2024dust3r} & 3.90 & 24.21 & 14.06 & 12.19 & 19.97 & 16.08 & 26.46 & 17.78 & 22.12 \\
MASt3R \cite{leroy2024grounding} & \snd 2.95 & 29.12 & 16.04 & \trd 8.47 & 16.12 & 12.30 & \trd 8.82 & 14.69 & 11.76 \\
SLAM3R \cite{liu2025slam3r} & 4.16 & \trd 10.17 & \trd 7.17 & \snd 7.10 & \snd 3.72 & \snd 5.41 & \snd 6.65 & \snd 3.36 & \snd 5.01 \\
MASt3R-SLAM \cite{murai2025mast3r} & 3.92 & 15.31 & 9.62 & 9.28 & 47.65 & 28.47 & 9.87 & 15.22 & 12.55 \\
VGGT-SLAM \cite{maggio2025vggt} & \trd 3.78 & \snd 9.30 & \snd 6.54 & 11.55 & \trd 9.15 & \trd 10.35 & 11.01 & \trd 5.66 & \trd 8.34 \\
\bf \method{} (ours) & \fst\textbf{2.89} & \fst\textbf{2.88} & \fst\textbf{2.89} & \fst\textbf{6.11} & \fst\textbf{1.86} & \fst\textbf{3.99} & \fst\textbf{4.26} & \fst\textbf{1.65} & \fst\textbf{2.96} \\ \bottomrule
\end{tabular}%
}
\end{table}

\begin{wraptable}{r}{0.45\textwidth}\vspace{-1cm}
\caption{\textbf{Tracking results on AriaMultiagent (multi-agent).}}
\label{tab:arial_tracking_multi}
\centering
\renewcommand{\tabcolsep}{8pt}
\resizebox{0.45\columnwidth}{!}{%
\begin{tabular}{lccc}
\toprule
\multirow{2}{*}{Methods} & \multicolumn{3}{c}{Multi-Agent}               \\ \cline{2-4} 
                         & Room 0        & Room 1        & Avg.          \\ \midrule
\multicolumn{4}{c}{RGB-D (Calibrated)}                                    \\ \midrule
Swarm-SLAM \cite{lajoie2023swarm} & 6.45          & 4.78          & 5.62          \\
CP-SLAM \cite{hu2023cp} & 3.03          & 2.87          & 2.95          \\
MAGiC-SLAM \cite{yugay2025magic}             & 1.15          & 0.65          & 0.90          \\ \midrule
\multicolumn{4}{c}{RGB (Uncalibrated)}                                   \\ \midrule
DUSt3R \cite{wang2024dust3r} $^\amalg$ & 15.46         & 21.36         & 18.41         \\
MASt3R \cite{leroy2024grounding} $^\amalg$ &  7.80           & 20.68         & 14.24         \\
SLAM3R \cite{liu2025slam3r} $^\amalg$ & \snd 6.97          & \snd 6.31          & \snd 6.64          \\
MASt3R-SLAM \cite{murai2025mast3r} $^\amalg$ & 9.13          & 29.34         & 19.24         \\
VGGT-SLAM \cite{maggio2025vggt} $^\amalg$                & 9.59          & \trd 8.47          & \trd 9.03          \\
MA-MASt3R-SLAM \cite{zhou2025multi}        & \trd 7.59          & 31.15         & 19.37         \\
\bf \method{} (ours)  & \fst \textbf{4.68} & \fst \textbf{2.36} & \fst \textbf{3.52} \\ \bottomrule
\end{tabular}%
}\vspace{-0.8cm}
\end{wraptable}

In Tab. \ref{tab:arial_tracking_multi}, we complete our evaluation by collecting results for the multi-agent setting. Once again, we report RGB-D systems at the top as a reference. We can appreciate how \method{} consistently outperforms any other multi-agent feed-forward method by a significant margin, highlighting, in particular, a notable gap with MA-MASt3R-SLAM \cite{zhou2025multi}. This further confirms the crucial role played by \ltog{}, outplaying merging strategies deployed by previous frameworks and concurrent multi-agent systems. 

\textbf{Supplementary material.} We refer the reader to the supplementary material for further experiments and insights.

\begin{wraptable}{l}{0.4\textwidth}\vspace{-1.3cm}
\centering
\caption{\textbf{Runtime analysis.} *taken from original paper.}
\label{tab:runtime}
\renewcommand{\tabcolsep}{10pt}
\resizebox{0.4\columnwidth}{!}{
\begin{tabular}{lc}
\toprule
\textbf{Method} & \textbf{FPS} \\ \midrule
CP-SLAM \cite{hu2023cp}* & $<1$ \\ 
MAGiC-SLAM \cite{yugay2025magic}* & $2$ \\ 
\midrule
DUSt3R \cite{wang2024dust3r} $^\amalg$ & $<1$ \\ 
MASt3R \cite{leroy2024grounding} $^\amalg$ & $<1$ \\ 
SLAM3R \cite{liu2025slam3r} $^\amalg$ & 3 \\ 
MASt3R-SLAM \cite{murai2025mast3r} $^\amalg$ & \trd 9 \\ 
VGGT-SLAM \cite{maggio2025vggt} $^\amalg$ & \fst 23 \\ 
MA-MASt3R-SLAM* \cite{zhou2025multi} & \snd 11 \\ 
\bf \method{} (ours) & \trd 9 \\ \bottomrule
\end{tabular}}\vspace{-0.8cm}
\end{wraptable}

\subsection{Runtime Comparison with RGB-D Systems}

In Tab. \ref{tab:runtime}, we conduct a runtime analysis on ReplicaMultiagent in the two-agent setup. RGB-D SLAM systems achieve higher tracking accuracy, yet existing solutions achieve very few FPS. In contrast, VGGT-SLAM \cite{maggio2025vggt} shows the full speed potential of feed-forward models, although falling short on accuracy.
Finally, both MA-MASt3R-SLAM \cite{zhou2025multi} and \method{} run at $\sim$10 FPS, although our framework retrieves significantly more accurate trajectories on AriaMultiagent.

\begin{figure}[h]
    \centering
    \includegraphics[width=0.98\linewidth]{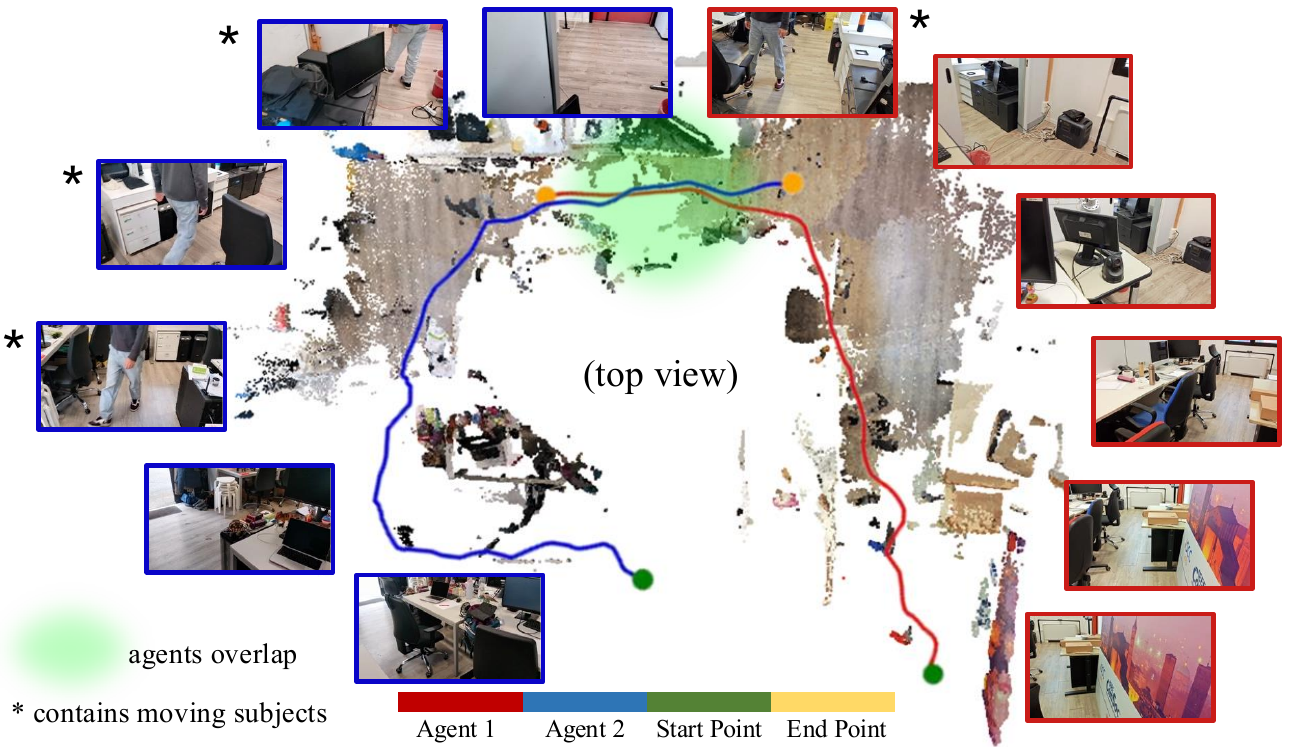} 
    \caption{\textbf{Experiments on a self-collected scene with limited overlap.} \method{} effectively works out globally consistent scene reconstruction.}\label{ref:custom} 
\end{figure}

\section{Qualitative results on custom sequences}\label{sec:qual}

We conclude with qualitative results on real-world video sequences, collected by 2 agents in low overlap ($\sim$15\%) and in the presence of some subjects moving in the scene, thus representing a particularly challenging scenario that pushes the limits of standard multi-agent reconstruction pipelines. Fig. \ref{ref:custom} shows the results by \method{}, looking qualitatively consistent despite such challenges.
\section{Conclusion}

In this paper, we presented the first multi-agent feed-forward RGB 3D reconstruction system, \method{}, achieving high-quality collaborative scene reconstruction and camera tracking in indoor scenes. Our feed-forward framework enables the system to reconstruct scenes at almost 10 FPS, with the proposed \ltog{} model enabling \method{} to merge the submaps from both intra-agent and inter-agent. Our experiments demonstrate that \method{} achieves both accurate reconstruction and camera tracking.

\textbf{Limitations and Future Work.} Our current framework, as well as the baselines, is not meant for agents facing very different conditions -- e.g., day vs night, adverse weather. 
We plan to extend this aspect, possibly with new benchmarks, as well as to evaluate multi-agent systems at a larger scale ($>3$ agents).


\bibliographystyle{splncs04}
\bibliography{main}
\end{document}